# An Improved Text Sentiment Classification Model Using TF-IDF and Next Word Negation

Bijoyan Das
Student Member, IEEE

Sarit Chakraborty
Member, IEEE, Kolkata, India

**Abstract –** With the rapid growth of Text sentiment analysis, the demand for automatic classification of electronic documents has increased by leaps and bound. The paradigm of text classification or text mining has been the subject of many research works in recent time. In this paper we propose a technique for text sentiment classification using term frequency- inverse document frequency (TF-IDF) along with Next Word Negation (NWN). We have also compared the performances of binary bag of words model, TF-IDF model and TF-IDF with 'next word negation' (TF-IDF-NWN) model for text classification. Our proposed model is then applied on three different text mining algorithms and we found the Linear Support vector machine (LSVM) is the most appropriate to work with our proposed model. The achieved results show significant increase in accuracy compared to earlier methods.

## I. INTRODUCTION

In recent past there has been a great hike in the usage of micro-blogging websites. These platforms have brought the entire world under a single domain, where everyone is free to share their opinions. The current decade has become a digital book where each post one shares cumulates to the sentiment of a particular topic. A lot of research works have been done to gather and calculate the sentiment of posts/tweets and also a good number of text-mining algorithms have been designed to analyze the sentiments.

The gradual growth of the number of users and the data related to them, has provided a good impetus to every company or organizations to mine these micro-blogging sites to collect information about people's opinion about their services or products. Due to this increase in user interaction, the future sales of any product or service depends a lot on the sentiments and perceptions of the previous buyers [1]. Therefore, it is necessary to have an efficient way to predict user sentiments about a product or service.

The solution to this problem is to classify the text using a strong machine-learning algorithm. Humans face many decisions on a daily basis and sentiment analysis can automate the process of coming to a decision based on past outcomes of that decision. For example, if someone has to buy tickets for a movie, then rather than manually going through all the long reviews, a sentiment classifier can predict the overall sentiment of the movie. Based on positive or negative sentiment a decision can be taken whether or not to buy tickets. Although this is a very trivial problem, text classification can be used in many different areas as follows:

- Most of the consumer based companies use sentiment classification to automatically generate reports on customer feedback. It is an integral part of CRM. [2]
- In medical science, text classification is used to analyze and categorize reports, clinical trials, hospital records etc. [2]
- Text classification models are also used in law on the various trial data, verdicts, court transcripts etc. [2]
- Text classification models can also be used for Spam email classification. [7]

In this paper we have demonstrated a study on the three different techniques to build models for text classification. The first two techniques which are simple binary bag of words model and TF-IDF model are common in text classification and we have tried to improve the accuracy by introducing next word negation. The performance of these techniques on several different machine learning algorithms is also shown at the end.

The methods to perform text classification can be broadly classified into supervised and unsupervised learning techniques [3].

The unsupervised learning techniques mainly use lexicon based approach where they use existing lexical resources like WordNet and language specific sentiment seed words to construct and update sentiment prediction [4]. Although unsupervised learning algorithms do not require a corpus of previously classified data and generates a general sentiment, they fail to capture context/domain specific information of the document.

The supervised learning techniques use machine learning on a previously classified dataset which is considered to be almost accurate. These pre-classified datasets are often domain specific, therefore the model it generate can work only for a particular domain. These datasets are first converted into intermediate models where documents are represented as vectors [6], and then the intermediate representations are fed to the machine learning algorithm. Through our research we have found out that Multinomial Naïve

Bayes, Max Entropy Random Forest and Linear Support Vector Machines are the popular choice of algorithm for text classification.

The documents are represented as a vector, where every word is converted into a number. This number can be binary (0 and 1) or it can be any real number in case of TF-IDF model. In case of binary bag of words model if a word appears in a document it gets a score 1 and if the word does not appear it gets a score 0. So, the document vector is a list of 1s and 0s. In case of TF-IDF the document vector can be a list of any numbers which are calculated using term frequency-inverse document frequency method.

In our work we have used three datasets, the IMDB movie review dataset [12], Amazon Product review dataset [5] and SMS Spam Collection dataset [8]. Each of these datasets have textual data pre-categorized into classes.

We have tried all the three approaches on these datasets starting with simple binary bag of words approach, then moving towards TF-IDF and TF-IDF with word negation approaches. In all the cases we have started with a base feature size and increased it gradually to produce better results.

In the next section we have displayed a survey of the various sentiment analysis techniques used all over the world, and then move towards our own proposed method, experiments and results.

## II. PRIOR WORK

Pang, Lee and Vaithyanathan were the first to propose sentiment classification using machine learning models. They analyzed the Naïve Bayes, Max Entropy and Support Vector Machine models for sentiment analysis on unigrams and bigrams of data [9]. In their experiment SVM paired with unigrams produced the best results.

Mullen and Collier performed sentiment classification using SVM by collecting data from a lot of sources [10]. Their work showed that using hybrid SVM with features based on Osgood's theory [ref.] produced the best results. This method worked well but failed to give importance to more contextual classifications and because of this domain variability the overall result greatly diminished. The accuracy rate from their proposed method was 86.6% which needs to be greatly improved.

Zhang constructed a computational model to explore reviews linguistics properties [14] to judge its usefulness. Support Vector Machine (SVM) algorithm was used for classification. Zhang concluded that the quality of review is good if it contains both subjective and objective information.

However, the efficiency of the analysis was only 72% because of employing fuzzy search technique to opinion mining which resulted in occurrence of a major problem when confronted by any misspelled word.

Efthymios et al. under-went sentiment analysis on Twitter messages using various features for classifications- N-gram feature, lexicon feature, POS feature. Their work was mainly subject specific and achieved an accuracy of nearly 80% and also concluded that POS feature diminishes accuracy level [16].

Farooq, et.al performed negation handling techniques in sentiment analysis [15]. They analyzed the effects of both syntactic and diminishing negation words in their experiment. They achieved an average accuracy rate of 83.3%.

## III. PROPOSED WORK

The overview of our proposed model is displayed in Figure 1. In the first phase we have imported the data from the specific domain and preprocessed that data removing the different punctuations.

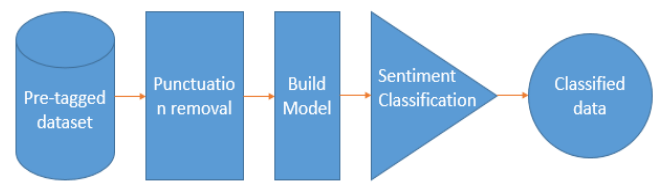

In the next stage the specific model is prepared using the preprocessed data. We have tested the performance of three different text representation models.

We have started with the simple binary bag of words of model where each document is represented as a fixed size vector of
0s and 1s where if a word appears in a document it gets a 1 and if it doesn't then it gets a 0. As an example, consider these four document below:

D1: the movie was a very indulging cinematic experience.
D2: standard of this movie is above its contemporaries.
D3: director brought out the best of the pair.
D4: moviegoers won't mind seeing the pair again.
The binary bag of words representation for these four documents using 8 frequently occurring words is shown in the table 1.

| Docs/ Words | the | movie | of | pair | was | a | wont | mind |
|---|---|---|---|---|---|---|---|---|
| D1 | 1 | 1 | 0 | 0 | 1 | 1 | 0 | 0 |
| D2 | 0 | 1 | 1 | 0 | 0 | 0 | 0 | 0 |
| D3 | 0 | 0 | 1 | 1 | 0 | 0 | 0 | 0 |
| D4 | 0 | 0 | 0 | 1 | 0 | 0 | 1 | 1 |

*Table 1. Binary Bag of Words Model*

This model is represents only the existence of words but does not take into account the importance of specific words in a document, like in the first document "indulging" is a much more important word compared to the other words for measuring the polarity of the sentence. But in this model all the words appearing in document 1 get the value '1' and words not appearing get a value of '0'. Thus, it is a binary model/two dimensional sentiment analysis model. This led us to try out some other enhanced bag of words models.

The second model we tested is the bag of words model with term frequency-inverse document frequency scores. Here the documents are also represented as vectors but instead of a vector of '0's and '1's, now the document contains scores for each of the words. These score are calculated by multiplying TF and IDF for specific words. So, the score of any word in any document can be represented as per the following equation:

$$TFIDF(word, doc) = TF(word, doc) * IDF(word)$$

Therefore in this method, two matrices have to be calculated, one containing the inverse document frequency of a word in the whole corpus of documents and another containing the term frequency of each word in each document. The formulae to calculate both of them are as follows:

$$TF(word, doc) = \frac{Frequency\ of\ word \in the\ doc}{No.\ of\ words \in the\ doc}$$

$$IDF(word) = log_e\left(1 + \frac{No.\ of\ docs}{No.\ of\ docs\ with\ word}\right)$$

The proposed example sentences can be converted into a TF-IDF model using the above method. Firstly, an IDF dictionary is created containing the 8 frequently occurring words and their IDF values. Then the TF dictionary is formulated containing the TF values for the corresponding words in each documents. The TFIDF model is shown in table 2.

This model is different compared to the simple binary bag of words model as it does not represent the documents as vectors of '0's and '1's, rather assigns more precise values within 0 and 1 [11].

The simple TF-IDF model works well and gives importance to the uncommon words rather than treating all the words as equal in case of binary bag of words model. This model however fails to perform accurately when it encounters any sentence containing negations. This negation is a very common linguistic construction that affects word/sentence polarity. Therefore, the model should be framed in such a way that if presence of negations is considered then better result can be obtained.

In the third model we have applied a negation strategy in which words are negated based on prior knowledge of polar expressions [13]. In this model whenever a negation word is tracked some changes are made to the words succeeding it. Many earlier proposed models [13] have also used this strategy before. Whenever a negation word is tracked all the words right after it are preceded with a 'not_' until a punctuation is received [13]. But this approach doesn't seem realistic to negate all the words as it will introduced a lot of unwanted words in the whole corpus.

We have modified this strategy and instead of negating all the words till punctuation, we have negated the **very next** word following the negation word. So, if the sentence,
*"The bird is not flying in the sky"* is received then instead of converting it to *"The bird is not_flying not_in not_the not_sky"*, we have converted it to *"The bird is not_flying in the sky"*. So, the job of tracking all the remaining punctuations after negation word is also excluded. In this approach meaning of the very next word followed by the negation is changed only and thus the meaning of the entire sentence gets changed.

| Docs/ Words | the | movie | of | pair | was | a | wont | mind |
|---|---|---|---|---|---|---|---|---|
| D1 | 0.09 | 0.09 | 0 | 0 | 0.17 | 0.17 | 0 | 0 |
| D2 | 0 | 0.09 | 0.09 | 0 | 0 | 0 | 0 | 0 |
| D3 | 0.18 | 0 | 0 | 0.09 | 0 | 0 | 0 | 0 |
| D4 | 0 | 0 | 0 | 0.09 | 0 | 0 | 0.17 | 0.17 |

*Table 2. TFIDF Model*

The algorithm for preprocessing in case of this model is shown in figure 1. The preprocessing phase of removing the punctuations, stop words is omitted in the algorithm and only the negation part is displayed to simplify the explanation. Our model therefore takes as input punctuation less documents. It then loops through the whole document and for each document performs the NWN technique.

Algorithm 1: Preprocessing of NWN

1.  Loop through the corpus of documents
2.      Set new_word_list to empty
3.      Convert document into lower case and remove single characters.
4.      Split each document into words.
5.      Loop through the list of words
6.         Check if negation word
7.         If negation word
              Set temp_word to "not"
8.         End
9.         Else if temp_word is "not"
10.           Add "not_" before the word
11.        End
12.        Append the word to the new_word_list
13.        Set temp_word to Empty
14.     End
15.     Join all words in the list to form a new document
16. End

*Figure 1. Proposed Algorithm for NWN*

After this preprocessing, a TFIDF model is formed in the same way as before. The proposed example sentences converted into the TFIDF NWN model is displayed in table 3.

It can be seen in this model above the word "not_mind" has higher score than both "wont" and "mind" in the simple TFIDF model.

After preparing the model for both training and testing using the text dataset, we have fitted the model in three popular classification algorithms as mentioned before i.e. Linear SVM [17], Multinomial Naïve Bayes [18] and Max Entropy Random Forest [19]. These models produces the various classifiers that can used to predict sentiment of new incoming data. In the next section the various experiments along with the obtained results are shown.

## IV.  EXPERIMENTAL RESULTS

We have used three different datasets and we have chosen the movie review dataset as the primary one for the experiments. For training and testing purposes we have split the dataset into two parts with split ratio of 0.8 (80% data for training and 20% data for testing). In the next sub sections we have conducted several experiments with different classification algorithms and also used various feature sizes.

| DATASET | SIZE / Nos. of Instances |
|---|---|
| **Movie Review Dataset** | 50,000 |
| **Product Review Dataset** | 50,000 |
| **SMS Spam Dataset** | 5,573 |

*Table 4. Data Volume*

### System Information:

Memory       : 8 GB DDR4 RAM
Speed         : 2.133 GHz
Processor    : Intel i5 8th Generation
Simulation    : Python 3.6

### 4.1 Exp. 1 (SVM with feature size from 2000-8000):

In the movie review dataset, we have used 40,000 reviews to train the classifier and 10,000 reviews to test the model performance. We have found out that accuracy of the classification increases as we increase the feature size. As shown in the graph below there is a steep increase in accuracy between the range of 2000-3000 and then it gradually slows down.

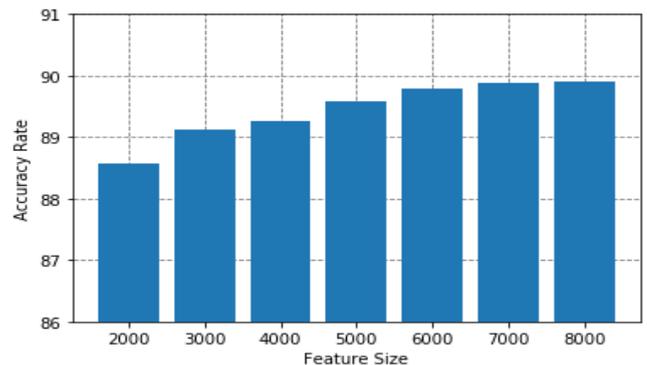

*Figure 2. Accuracy Rate vs Feature Size*

### 4.2 Exp. 2 (10-Fold Cross Validation with SVM):

After testing the accuracy of the model on the test set, we have used 10 fold cross validation technique to find out the average accuracy. Dataset was divided in 10 parts. At every fold 90% of the data are used to train the model and 10% of the data are used to check model performance. Therefore, after the process we get a list of accuracies of the different folds. So, we can calculate the mean of these different accuracies to better understand the accuracy of the model. Fig. 3 below shows the accuracies obtained in each fold of the training set with 5000 features for 10 fold cross validation.

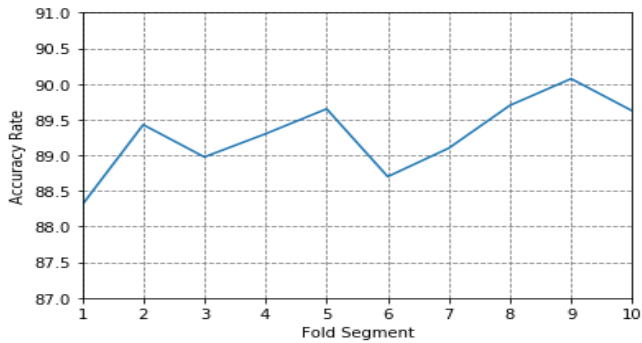

*Figure 3. Accuracy Rate vs Fold Segment*

### 4.3 Exp. 3 (Accuracy comparison of three algorithms):

We have used three of the most popular algorithms to train the classifier. The accuracy rate from each of the different algorithms for 10,000 features of data are displayed in the table 5 below. We found out that Linear Support Vector Machine produces the highest amount of accuracy among the three.

| Algorithm | Accuracy |
|-----------|----------|
| Linear Support Vector Machine (LSVM) | 89.91 |
| Multinomial Naïve Bayes (MVB) | 86.34 |
| Max Entropy Random Forest (MERF) | 86.08 |

*Table 5. Accuracies in ML Algorithms*

### 4.4 Exp. 4 (10-Fold Cross Validation Comparison):

In this experiment we have tested and compared the 10-Fold cross validation results for each of the algorithm for IMDB movie review dataset as follows. The figure 4 below displays the 10-Fold cross validation results for each of the algorithms. These results are from testing with 5000 features.

{ Explain how Graph-→ goes on }

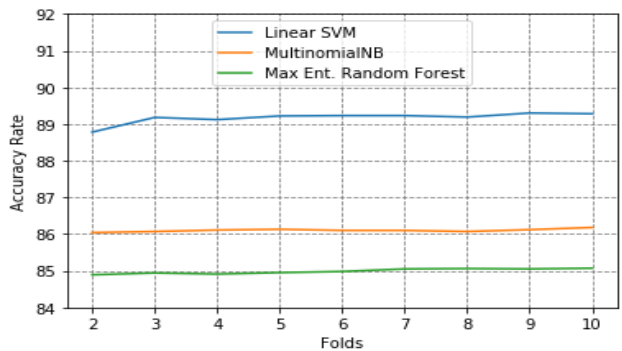

*Figure 4. 10-Fold Cross Validation Comparison*

### 4.5 Exp. 5 (Performance in different Datasets):

Apart from using the IMDB large movie review dataset [12] we have also tested the model performance on two other datasets from two different domain. While measuring the accuracies of each of our model in each of these datasets we have taken 8000 features into account. The accuracies from the different datasets are displayed in the table 6 below.

| Dataset | Accuracy |
|---------|----------|
| IMDB Movie Review Dataset | **89.91%** |
| SMS Spam Collection Dataset | **96.83%** |
| Amazon Product Review Dataset | **88.58%** |

*Table 6. Accuracies in Datasets*

### 4.6 Exp. 6 (Final Comparison of Different Models):

In this experiment we have compared the performances of the three text representation models on the movie review dataset. In this experiment 8000 features are taken into consideration for maximum output accuracy. The accuracies for these models are displayed in the figure 5 below and it can be seen that our model (TFIDF with next word negation) outperforms both the basic binary bag of words representation and the simple TFIDF representation.

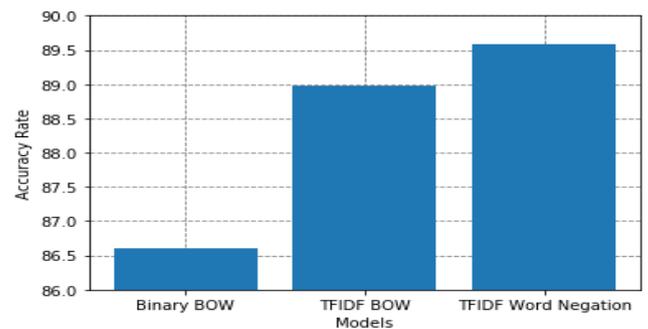

*Figure 5. Model Performance Comparison*

## V. CONCLUSION

In this paper, we have conducted experiments on three datasets, IMDB movie reviews [12], Amazon product reviews [5] and SMS spam detection dataset [8]. After performing sentiment analysis on these datasets using binary bag of words model and TF-IDF model we found out the accuracy as XX.XX and XX.XX respectively [Fig. 5]. But, after conducting experiment with our proposed model, i.e, using NWN with TF-IDF, we found out a good increase in the accuracy level. The accuracy percentages for IMDB review datasets, Amazon product review and SMS spam datasets came as 89.91%, 88.86%, 96.83% respectively[Table 6].

So, from our experiments, we have concluded that when TF-IDF model is coupled with Next Word Negation then the performance of the sentiment classifier increases by a good percentage.

| Docs/Words | the | movie | of | pair | was | a | not_mind | cinematic |
|---|---|---|---|---|---|---|---|---|
| D1 | 0.09 | 0.09 | 0 | 0 | 0.17 | 0.17 | 0 | 0.17 |
| D2 | 0 | 0.09 | 0.09 | 0 | 0 | 0 | 0 | 0 |
| D3 | 0.18 | 0 | 0 | 0.09 | 0 | 0 | 0 | 0 |
| D4 | 0 | 0 | 0 | 0.10 | 0 | 0 | 0.19 | 0 |

*Table 3. TFIDF Model with NWN*

In future we seek to further improve the accuracy of this model by working on contextual opposite of the word following the negation word.